\documentclass[10pt,twocolumn,letterpaper]{article}

\usepackage{iccv}
\usepackage{times}
\usepackage{epsfig}
\usepackage{graphicx}
\usepackage{amsmath}
\usepackage{amssymb}

\usepackage{booktabs}
\usepackage{siunitx}
\usepackage{multirow}
\usepackage{multicol}
\usepackage{color}
\usepackage{pifont}
\usepackage{tabularx}
\usepackage{xpatch}
\usepackage{floatrow}
\usepackage{bbm}
\usepackage[dvipsnames]{xcolor}
\usepackage{arydshln}

\usepackage[pagebackref=true,breaklinks=true,letterpaper=true,colorlinks,bookmarks=false]{hyperref}

\iccvfinalcopy 

\def\iccvPaperID{3114} 

\ificcvfinal\fi

\begin{document}

\title{Instance-level Image Retrieval using Reranking Transformers}

\author{
Fuwen Tan \\
University of Virginia\\
{\tt\small fuwen.tan@virginia.edu}
\and
  Jiangbo Yuan \\
  eBay Computer Vision\\
    {\tt\small jiayuan@ebay.com}
\and
  Vicente Ordonez \\
  Rice University \\
    {\tt\small vicenteor@rice.edu}
}

\maketitle
\ificcvfinal\thispagestyle{empty}\fi

\begin{abstract}
Instance-level image retrieval is the task of searching in a large database for images that match an object in a query image. To address this task, systems usually rely on a retrieval step that uses global image descriptors, and a subsequent step that performs domain-specific refinements or reranking by leveraging operations such as geometric verification based on local features. In this work, we propose Reranking Transformers (RRTs) as a general model to incorporate both local and global features to rerank the matching images in a supervised fashion and thus replace the relatively expensive process of geometric verification. RRTs are lightweight and can be easily parallelized so that reranking a set of top matching results can be performed in a single forward-pass. We perform extensive experiments on the Revisited Oxford and Paris datasets, and the Google Landmarks v2 dataset, showing that RRTs outperform previous reranking approaches while using much fewer local descriptors.
Moreover, we demonstrate that, unlike existing approaches, RRTs can be optimized jointly with the feature extractor, which can lead to feature representations tailored to downstream tasks and further accuracy improvements. 
The code and trained models are publicly available at
\footnotesize{\href{https://github.com/uvavision/RerankingTransformer}{\texttt{github.com/uvavision/RerankingTransformer}}}.
\end{abstract}

\section{Introduction}
Instance recognition is a challenging task that aims to visually recognize an object instance. 
This is distinct from category-level recognition that identifies only the object class. 
Instance recognition is important in e-commerce where it is often desired to find a specific product in a large image collection, or in place identification where the objective is to infer the identity of a public landmark. 
As the number of instances is much larger than the number of object categories, instance recognition is typically cast as image retrieval instead of classification, and usually involves both metric learning and local feature based reranking.

\begin{figure}[t]
  \centering
  \includegraphics[width=0.8\textwidth]{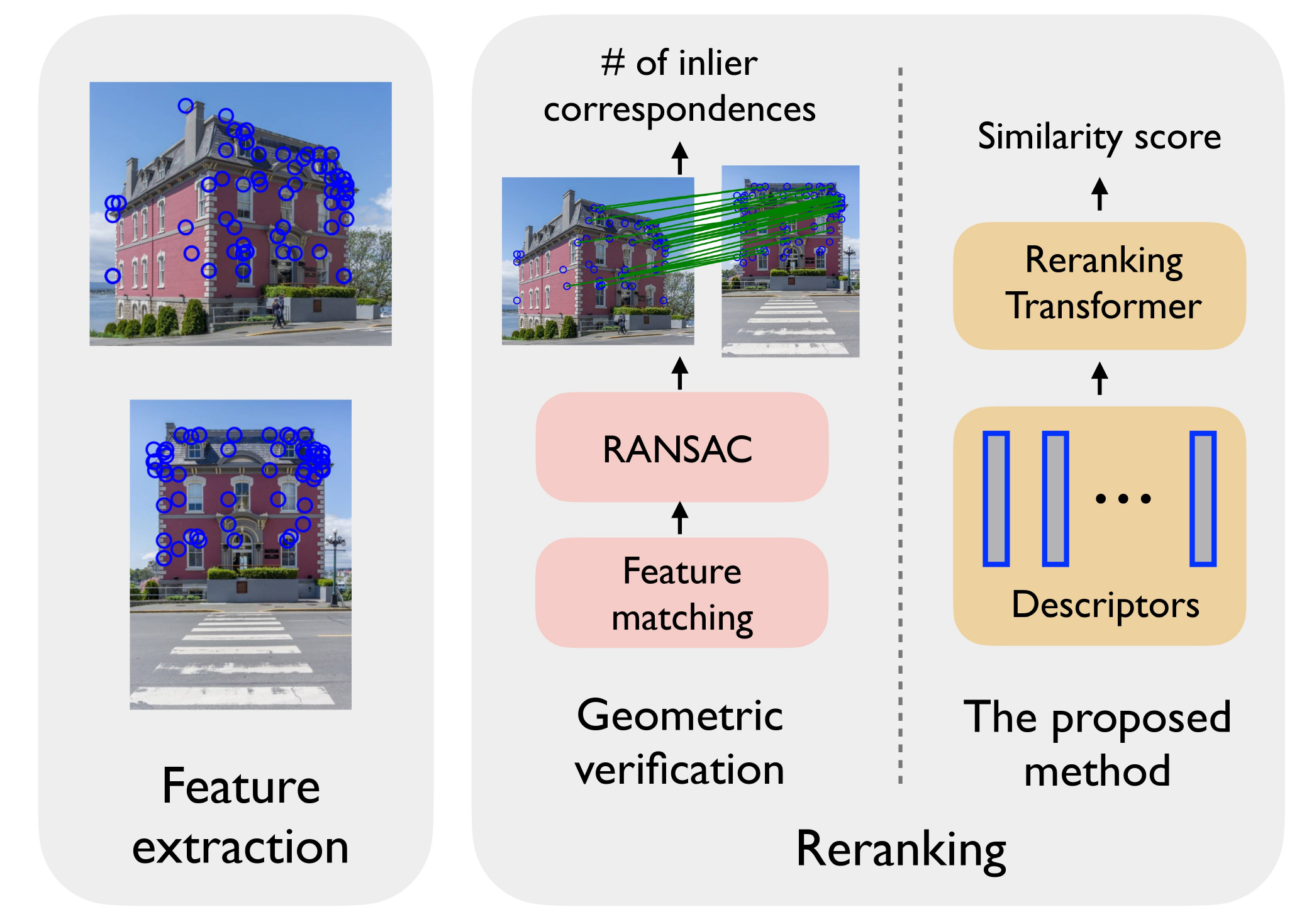}
  \vspace{-0.1in}
  \caption{
  Top performing instance recognition methods often rely on reranking the top results using a score such as the number of inlier correspondences from geometric verification. We propose to replace this step with a Reranking Transformer (RRT) that can be learned with the underlying representations of the images.
  \vspace{-0.25in}
  }
  \label{fig:teaser}
\end{figure}

Over the last decade, instance recognition continues to be a major focus of research. 
Pioneering systems leveraged hand-crafted local descriptors and matching algorithms~\cite{videogoogle,spatial2007}.
More recent approaches incorporate both global and local descriptors extracted from deep learning models~\cite{neuralcodes2014,delf2017}. Global descriptors summarize an image into a single vector, leading to a compact representation for large-scale search. 
Local descriptors encode detailed spatial features for patch-level matching, and are shown to be important for high retrieval precision~\cite{how2020,delg2020}. 
The best methods typically use a global descriptor to reduce the solution space to a set of candidate matching images, and local descriptors to \textit{re-rank} the nearest images~\cite{dsm2019, delg2020}.
While extensive progress has been made to improve image retrieval using global features, fewer efforts have been made to develop similarity metrics based on local features.
State-of-the-art approaches still rely on classic matching techniques, such as geometric verification~\cite{spatial2007} and aggregated selective match kernels (ASMK)~\cite{asmk2016}. 
Geometric verification assumes object instances are rigid and local matches between images can be estimated as a homography using RANSAC~\cite{ransac1981}. It is also an expensive process that requires iterative optimization on a large set of local descriptors.
ASMK aggregates the similarities of features without modeling the geometric alignment, but requires offline clustering and encoding procedures. 
It was mainly used as a global retrieval technique in previous literature.
Both geometric verification and ASMK require large amounts of local descriptors to ensure retrieval performance.


In this work, we propose \textit{Reranking Transformers} (RRTs), which learn to predict the similarity of an image pair directly. 
Our method is general and can be used as a drop-in replacement for other reranking approaches such as geometric verification. 
We conduct detailed experiments showing that as either a drop-in replacement or trained together with a global retrieval approach, the proposed method is the top-performing across the standard benchmarks for instance recognition. 
RRTs leverage the transformer architecture~\cite{transformer2017} which has led to significant improvements in natural language processing~\cite{bert2019,roberta2019} and vision-and-language tasks~\cite{visualbert2019,chen2020uniter,ViLBERT2019}.
Most recently, it has also been used for purely vision tasks, notably for image recognition~\cite{vision_transformer2021} and object detection~\cite{detr2020}. 
To the best of our knowledge, our work is the first to adapt transformers for a visual task involving the analysis of image pairs in the context of reranking image search results. 


Reranking Transformers are lightweight.
Compared with typical feature extractors which have over 20 million parameters (e.g.~25 million in ResNet50~\cite{resnet2016}), the proposed model only has 2.2 million parameters.
It can also be easily parallelized such that re-ranking the top 100 neighbors requires a single pass.
As shown in Fig.~\ref{fig:teaser}, our method directly predicts a similarity score for the matching images, instead of estimating a homography, which may be challenging under large viewpoint changes or infeasible for deformable objects.
Our method requires much fewer descriptors but achieves superior performance, especially for challenging cases.
In current state-of-the-art models, the feature extraction and matching are optimized separately, which may lead to suboptimal feature representations.
In this work, we first perform experiments using pretrained feature extractors, then demonstrate the benefit of jointly optimizing the feature extractor and our model in a unified framework. 

\textbf{Contributions}. 
(1) We propose \textit{Reranking Transformers} (RRTs), a small and effective model which learns to predict the similarity of an image pair based on global and local descriptors; 
(2) Compared with existing methods, RRTs require fewer local descriptors and can be parallelized so that reranking the top neighbors only requires a single forward-pass; 
(3) We perform extensive experiments on Revisited Oxford/Paris~\cite{revisited}, Google Landmarks v2~\cite{gldv2}, and Stanford Online Products, and show that RRTs outperform prior reranking methods across a variety of settings.
\section{Related Work}
\begin{figure*}[t]
  \centering
  \includegraphics[width=0.8\textwidth]{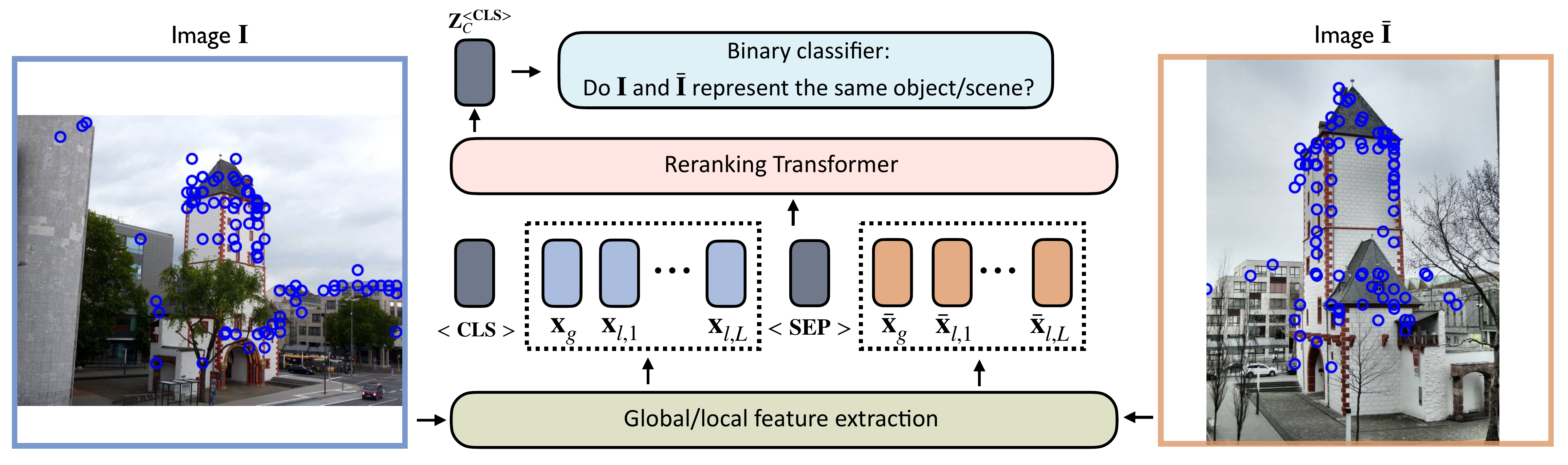}
  \vspace{-0.08in}
  \caption{Illustration of the proposed \textit{Reranking Transformer} (RRT) model. The input of RRT is a sequence of global and local descriptors (circled in {\color{blue}blue}) extracted from an image pair $(\mathbf{I}, \bar{\mathbf{I}})$). This sequence, together with two special tokens, are fed into a multi-layer transformer model which produces a similarity score of $(\mathbf{I}, \bar{\mathbf{I}})$. The model is trained to optimize a binary cross entropy loss.
  \vspace{-0.12in}
  }
  \label{fig:model}
\end{figure*}

\noindent \textbf{Feature learning for instance recognition}.
Hand-crafted local descriptors~\cite{local2005, SIFT2004} were widely used in earlier instance retrieval work~\cite{videogoogle, voctree2006}. 
Recently, local features extracted from convolution neural networks (CNN) are shown to be more effective~\cite{delf2017, rasmk2019, dsm2019, d2net2019,shanmin2018,pang2018building}. 
Some of these works learn feature detection and representation jointly by non-local maximum suppression~\cite{d2net2019,how2020}, or attention~\cite{delf2017, rasmk2019, delg2020}. 
The detected local descriptors are usually used for geometric verification~\cite{spatial2007} or ASMK~\cite{asmk2016}. 
Compared to local features, global descriptors provide a compact representation of an image for large-scale search. 
Current global descriptors are typically extracted from CNN models~\cite{neuralcodes2014, rmac2016, finetune2016, DIR2017} by 
spatial pooling~\cite{spoc2015, cdw2015, rmac2016, finetune2016,  solar2020}, which may not be ideal for modeling region-wise relations across images. 
Recent systems either use global descriptors to reduce the solution space and then local descriptors to \textit{re-rank} the nearest neighbors, or encode local descriptors using a large visual codebook, followed by image matching with an aggregated selective match kernel~\cite{asmk2016,rasmk2019, how2020}. 
This work mainly follows the retrieve-and-rerank paradigm.

\noindent\textbf{Reranking for instance recognition/retrieval.}
Geometric verification is the dominant image reranking approach and widely used in both traditional~\cite{spatial2007} and more recent works~\cite{dsm2019, delf2017, delg2020}.
Inspired by text retrieval, query expansion techniques have also been introduced for image retrieval~\cite{qe2007, qe2011, qe2014, attentionQE2020}. 
These methods differ from geometric verification and our work as they rely on analysing the local nearest neighbor graph for each query during testing. 
Diffusion based approaches~\cite{diffusion2013, rankfusion2015, diffusion2017, Bai_2019_CVPR, diffusion4} aim to learn the structure of the data manifold by similarity propagation over the global affinity graph built on a query and all the gallery images, which is nontrivial to scale.
Overall, the motivation of image reranking is to make better use of test-time knowledge to refine retrieval results. 
Our work shares the same vision with this line of research but focuses more on learning the similarity of an image pair directly.

\noindent\textbf{Transformers for visual tasks.}
Transformers have become the dominant architecture for representing text~\cite{bert2019, roberta2019}. 
Recently, it has also been introduced to vision-and-language~\cite{visualbert2019, ViLBERT2019} and pure vision tasks~\cite{image_transformer2018, detr2020,lanchantin2021general}. 
As the key ingredient of the transformer architecture, the self-attention mechanism has also been studied for visual recognition~\cite{aacnn2019,standalone2019,san2020}.
These works apply transformers for single image predictions while we leverage transformers to learn the visual relation of an image pair. Our work is also closely related to SuperGlue~\cite{superglue2020}, which while not a Transformer, also relies on self-attention. SuperGlue aims to learn local correspondences between images with pixel-level supervision. Our work differs from SuperGlue in that it learns the similarity of an image pair with image-level supervision. We provide a best-effort comparison with this approach.
\section{Methodology}

\subsection{Attention Modules in Transformers}
\label{sec:attention}
First, we briefly review the key ingredients in the Transformer architecture: Single-Head Attention (SHA) and Multi-Head Attention (MHA).

\textbf{Single-Head Attention (SHA)}:
The input of a SHA layer comprises three sets of variables: the queries $\textsc{Q} := \{\mathbf{q}_i \in \mathbb{R}^{d_q}\}_{i=1}^N$, the keys $\textsc{K} := \{\mathbf{k}_j \in \mathbb{R}^{d_k}\}_{j=1}^M$, and the values $\textsc{V} := \{\mathbf{v}_j \in \mathbb{R}^{d_v}\}_{j=1}^M$. 
Here, $d_q$, $d_k$, $d_v$ are the dimensions of the corresponding feature vectors, $N$ and $M$ are the sequence lengths. 
SHA produces a new feature sequence where each vector is a linear combination of $\{\mathbf{v}_j\}$. 
In doing this, Q, K, V are first linearly projected as $\bar{\textsc{Q}} = \textsc{Q}W^Q,~\bar{\textsc{K}} = \textsc{K}W^K,~\bar{\textsc{V}} = \textsc{V}W^V$, using parameter tensors: $W^Q \in \mathbb{R}^{d_q \times d_h},~W^K \in \mathbb{R}^{d_k \times d_h},~W^V \in \mathbb{R}^{d_v \times d_h}$, where $d_h$ is the new feature dimension. 
The output of a SHA layer is computed as: $\textsc{SHA(Q, K, V)} := \textsc{softmax}(\frac{\bar{\textsc{Q}}\bar{\textsc{K}}^T}{\sqrt{d_h}})\bar{\textsc{V}}$.

\textbf{Multi-Head Attention (MHA)}:
Like SHA, MHA takes Q, K, V as input and comprises multiple SHA modules: $\textsc{MHA(Q, K, V)} := [\textsc{head}_1;\cdots;\textsc{head}_h]W^O$, $\textsc{head}_i := \textsc{SHA}_i\textsc{(Q, K, V)}$. 
Here $[;]$ denotes the concatenation operator, $h$ is the number of the SHA heads. $W^O \in \mathbb{R}^{(hd_h) \times d_e}$ is a linear projection with an output dimension of $d_e$.

\subsection{Model}
With the fundamental building blocks defined above, we introduce the detailed formulation of our model:

\textbf{Image representations}: An image $\mathbf{I}$ is represented by a global descriptor of a dimension $d_g$: $\mathbf{x}_g \in \mathbb{R}^{d_g}$ and a set of $\textit{L}$ local descriptors: $\mathbf{x}_l = \{\mathbf{x}_{l, i} \in \mathbb{R}^{d_l}\}_{i=1}^L$, each of a dimension $d_l$. 
Both $\mathbf{x}_g$ and $\mathbf{x}_l$ are extracted from a CNN backbone (to be discussed in Sec.~\ref{sec:impl}).
Optionally, each $\mathbf{x}_{l, i}$ is associated with a coordinate tuple $\mathbf{p}_{l, i} = (u, v) \in \mathbb{R}^2$ and a scale factor $s_{l, i} \in \mathbb{R}$, indicating the pixel location and image scale where $\mathbf{x}_{l, i}$ is extracted from. 
In this work, $s_{l, i}$ is an integer, indexing a set of pre-defined image scales.

\textbf{Input}:
As a sequence transduction model~\cite{bert2019, roberta2019}, Transformers take as input a list of ``tokens'' (e.g. $\textsc{Q}$,$\textsc{K}$,$\textsc{V}$ in Sec.~\ref{sec:attention}).
In image retrieval, these ``tokens'' can be derived from the features of an image pair $(\mathbf{I}, \bar{\mathbf{I}})$. Following the BERT transformer encoder~\cite{bert2019}, we define the input as:
\begin{equation}
\begin{aligned}
\mathbf{X}(\mathbf{I}, \bar{\mathbf{I}}) := [~&\langle\textsc{CLS}\rangle; f_g(\mathbf{x}_g); f_l(\mathbf{x}_{l,1}); \cdots ; f_l(\mathbf{x}_{l,L}); \\
              &\langle\textsc{SEP}\rangle; \bar{f}_g(\mathbf{\bar{x}}_g); \bar{f}_l(\mathbf{\bar{x}}_{l,1});\cdots ;\bar{f}_l(\mathbf{\bar{x}}_{l,L})~],
\label{equ:input}
\end{aligned}
\end{equation}
where:
\begin{equation}
\begin{aligned}
f_g(\mathbf{x}_g) &:= \mathbf{x}_g + \alpha;\\
f_l(\mathbf{x}_{l,i}) &:= \mathbf{x}_{l, i} + \varphi(\mathbf{p}_{l, i}) + \psi(s_{l, i}) + \beta\\
\bar{f}_g(\mathbf{\bar{x}}_g) &:= \mathbf{\bar{x}}_g + \bar{\alpha};\\
\bar{f}_l(\mathbf{\bar{x}}_{l,i}) &:= \mathbf{\bar{x}}_{l,i} + \varphi(\mathbf{\bar{p}}_{l, i}) + \psi(\bar{s}_{l, i}) + \bar{\beta}.
\label{equ:embed}
\end{aligned}
\end{equation}

Here, $\langle\textsc{CLS}\rangle$ is a special token used for summarizing the signals from both images. $\langle\textsc{SEP}\rangle$ is an extra separator token. 
$\alpha, \bar{\alpha}, \beta, \bar{\beta}$ are one dimensional segment embeddings, being used to distinguish the global and local descriptors of $\mathbf{I}$ and $\bar{\mathbf{I}}$. $\varphi$ is a position embedding function, as used in~\cite{detr2020}.
$\psi$ is a linear embedding taking the scale index $s_{l, i}$ as input.

\textbf{Model architecture}:
With input $\mathbf{X}(\mathbf{I}, \bar{\mathbf{I}})$, we define a multi-layer transformer where each layer is formulated as:
\begin{equation}
\begin{aligned}
\mathbf{\bar{Z}}_{i+1} &=\textsc{LayerNorm}(\mathbf{Z}_i + \textsc{MHA}(\mathbf{Z}_i)),\\
\mathbf{Z}_{i+1} &= \textsc{LayerNorm}(\textsc{MLP}(\mathbf{\bar{Z}}_{i+1})),~~~~~~\\
\textsc{MLP}(\mathbf{\bar{Z}}_{i+1}) &= \textsc{ReLU}(\mathbf{\bar{Z}}_{i+1}W_1^T)W_2^T,~~~~~~~~~~~~~~~~\\
i &= 0, \cdots, C-1.
\label{equ:trans}
\end{aligned}
\end{equation}

In this setting, the Q, K, V features for MHA are the same set of vectors $\textbf{Z}_i$, with $\mathbf{Z}_0 = \mathbf{X}(\mathbf{I}, \bar{\mathbf{I}})$.
$\textsc{MLP}$ is a two-layer perceptron with parameter matrices $W_1 \in \mathbb{R}^{d_e \times d_c}$ and $W_2 \in \mathbb{R}^{d_c \times d_e}$, and an intermediate dimension $d_c$. $\textsc{LayerNorm}$ is a layer normalization function proposed in~\cite{layernorm2016}. The model includes \textit{C} transformer layers in total.

\textbf{Training objective}:
Our model is trained to optimize a binary cross entropy loss:
\begin{eqnarray}
\textsc{E}(\mathbf{I}, \bar{\mathbf{I}}) = \textsc{BCE}(\textsc{sigmoid}(\mathbf{Z}_C^{\langle\textsc{CLS}\rangle} W_z^T), \mathbbm{1}(\mathbf{I}, \bar{\mathbf{I}})),
\label{equ:bce}
\end{eqnarray}
$\mathbf{Z}_C^{\langle\textsc{CLS}\rangle} \in \mathbb{R}^{d_e}$ is a feature vector, corresponding to the $\langle\textsc{CLS}\rangle$ token.
It is extracted from the last transformer layer.
$W_z^T \in \mathbb{R}^{d_e \times 1}$ is a linear function mapping $\mathbf{Z}_C^{\langle\textsc{CLS}\rangle}$ to a logit scalar. 
$\mathbbm{1}(\mathbf{I}, \bar{\mathbf{I}})$ is an indicator function which equals to one when $\mathbf{I}$ and $\bar{\mathbf{I}}$ represent the same object, or zero otherwise. Fig.~\ref{fig:model} provides an illustration of the proposed model.

\section{Experiments}
\vspace{-0.05in}
Next, we describe the datasets we use to evaluate our approach, and details about our implementation.
\vspace{-0.05in}

\subsection{Datasets}
We perform experiments on three datasets, Google Landmarks v2 (GLDv2)~\cite{gldv2}, Revisited Oxford/Paris~\cite{revisited}, and Stanford Online Products (SOP)~\cite{sop2016}. 

\textbf{GLDv2}: Google Landmarks v2 (GLDv2) ~\cite{gldv2} is a new benchmark for instance recognition that includes over five million images from 200k natural landmarks.
As the Reranking Transformer has limited parameters (e.g. 2.2 million), we sample a small subset of the images from the ``v2-clean'' split of GLDv2 for training.
We \textit{randomly} sample 12,000 landmarks where each landmark has \textit{at least} 10 images.
For each landmark, we \textit{randomly} sample \textit{at most} 500 images.
This results in 322,008 images, which is 20\% of the ``v2-clean'' split and 8\% of the original training set. 
The names of the sampled images are included in the supplementary material.
For testing, we evaluate on the standard test set for the retrieval task, which contains 1,129 query images and 761,757 gallery images.

\textbf{$\mathcal{R}$Oxf} and \textbf{$\mathcal{R}$Par}: 
Revisited Oxford ($\mathcal{R}$Oxf) and Paris ($\mathcal{R}$Par)~\cite{revisited} are standard benchmarks for instance recognition, which have 4,993 and 6,322 gallery images respectively. 
They both have 70 query images, each with a bounding box depicting the location and span of the prominent landmark. 
An extra distractor set ($\mathcal{R}$1M) with 1,001,001 images is included for large-scale experiments.
We follow the standard evaluation protocol~\cite{revisited, delg2020} and crop the query image using the provided bounding box.
We report mean Average Precision (mAP) on the Medium and Hard setups.

\textbf{SOP}: To investigate the benefit of jointly optimizing the feature representation and our Reranking Transformer, we perform experiments on a dataset of product images: Stanford Online Products (SOP)~\cite{sop2016}. 
SOP is a commonly used benchmark for metric learning~\cite{margin2017,divide2019,MIC2019, fastap2019,roth2020revisiting, XBM2020, CE2020}, which includes 120,053 images, 59,551 for training, 60,502 for testing. 
We follow the evaluation protocol for metric learning and report the R@K scores.

\subsection{Implementation}
\label{sec:impl}
\textbf{Experiments on the pretrained descriptors}: We first perform experiments with descriptors obtained from a pretrained feature extractor, DELG~\cite{delg2020}.
Our main experiments leverage ResNet50~\cite{resnet2016} as the CNN backbone, but we also include experiments with ResNet101 in the supplementary material.
DELG provides a unified framework for global/local feature extraction. 
The local descriptors, each with a dimension of 128, are extracted at 7 image scales ranging from 0.25 to 2.0.
The global descriptor with a dimension of 2048 is extracted at 3 scales: \{$\frac{1}{\sqrt{2}}$, 1, $\sqrt{2}$\}.
We use an extra linear projection to reduce the global descriptor to a dimension of 128. 
In the original DELG model, the top 1000 local descriptors with the highest attention scores are selected for image reranking.
We observe that RRT does not require this amount of descriptors, and the retrieval performance saturates at 500 local descriptors.
Thus, in our experiments we choose the top 500 local descriptors and set $L=500$, $d_g = d_l = 128$.
For images with fewer descriptors, we pad the feature sequence with empty vectors and use a binary attention mask, as in BERT~\cite{bert2019}, to indicate the padding locations.
Both the global and local features are L2 normalized to unit norm.
During training, the positive image is \textit{randomly} sampled from the images sharing the same label as the query.
The negative image is \textit{randomly} sampled from the top 100 neighbors returned by the global retrieval, which have a different label from the query.
DELG is pretrained on both Google Landmarks (GLD) v1~\cite{delf2017} and v2-clean~\cite{gldv2}. Thus, we perform experiments on two sets of descriptors extracted from these two models. For the architecture, we use 4 SHA heads ($h=4$) and 6 transformer layers ($C=6$). $d_q$, $d_k$, $d_v$ and $d_e$ in SHA are set to 128, $d_h$ is set to 32, $d_c$ in \textsc{MLP} (Eq.~\ref{equ:trans}) is set to 1024. 
The number of learnable parameters is 2,243,201, which is 9\% of the amount in ResNet50.
The model is trained with AdamW~\cite{adamw2019} for 15 epochs, using a learning rate of 0.0001 and a weight decay of 0.0004.

\begin{table*}[t]
\scalebox{0.94}{
\setlength{\tabcolsep}{5pt}
\centering
{
\begin{tabular}{l l l l c c c c c c c c c}
\toprule
\multirow{2}{*}{Method} & \# local & \# Reranked & Desc.  & \multicolumn{4}{c}{ Medium} && \multicolumn{4}{c}{ Hard} \\
& desc. & images & version & \multicolumn{1}{c}{\normalsize \vphantom{M} \normalsize $\mathcal{R}$Oxf } & \multicolumn{1}{c}{\normalsize +$\mathcal{R}$1M} & \multicolumn{1}{c}{\normalsize $\mathcal{R}$Par} & \multicolumn{1}{c}{\normalsize +$\mathcal{R}$1M} && \multicolumn{1}{c}{\normalsize $\mathcal{R}$Oxf} & \multicolumn{1}{c}{\normalsize +$\mathcal{R}$1M} & \multicolumn{1}{c}{\normalsize $\mathcal{R}$Par} & \multicolumn{1}{c}{\normalsize +$\mathcal{R}$1M} \\
\midrule
DELG global & 0 & 0& R50-v1 & $69.7$ & $55.0$ & $81.6$ & $59.7$ && $45.1$ &$27.8$ & $63.4$ & $34.1$ \\
\midrule
GV & 1000 & 100 & R50-v1 & $75.4$ &$61.1$ & $82.3$ & $60.5$ && $54.2$ & $36.8$ & $64.9$ & $34.8$ \\ 
RRT (ours) & 500 & 100 & R50-v1 & $\num[math-rm=\mathbf]{75.5}$ &$\num[math-rm=\mathbf]{61.2}$ & $\num[math-rm=\mathbf]{82.7}$ & $\num[math-rm=\mathbf]{60.7}$ && $\num[math-rm=\mathbf]{56.4}$ & $\num[math-rm=\mathbf]{37.0}$ & $\num[math-rm=\mathbf]{68.6}$ & $\num[math-rm=\mathbf]{37.5}$ \\
\midrule
GV$\ast$ & 1000 & 200 & R50-v1 & $77.2$ &$63.1$ & $82.5$ & $60.9$ && $55.4$ & $37.9$ & $63.2$ & $34.7$ \\ 
RRT (ours) & 500 & 200 & R50-v1 & $\num[math-rm=\mathbf]{77.9}$ &$\num[math-rm=\mathbf]{63.5}$ & $\num[math-rm=\mathbf]{84.4}$ & $\num[math-rm=\mathbf]{62.1}$ && $\num[math-rm=\mathbf]{58.8}$ & $\num[math-rm=\mathbf]{39.5}$ & $\num[math-rm=\mathbf]{71.6}$ & $\num[math-rm=\mathbf]{39.5}$ \\ 
\midrule
DELG global &0 & 0& R50-v2-clean & $73.6$ & $60.6$ & $85.7$ & $68.6$ && $51.0$ & $32.7$ & $71.5$ & $44.4$ \\
\midrule
GV & 1000 & 100 & R50-v2-clean & $\num[math-rm=\mathbf]{78.3}$ & $\num[math-rm=\mathbf]{67.2}$ & $85.7$ & $69.6$ && $57.9$ & $43.6$ & $71.0$ & $45.7$ \\
RRT (ours) & 500 & 100 & R50-v2-clean & $78.1$ & $67.0$ & $\num[math-rm=\mathbf]{86.7}$ & $\num[math-rm=\mathbf]{69.8}$ && $\num[math-rm=\mathbf]{60.2}$ & $\num[math-rm=\mathbf]{44.1}$ & $\num[math-rm=\mathbf]{75.1}$ & $\num[math-rm=\mathbf]{49.4}$ \\
\midrule
GV$\ast$ & 1000 & 200  & R50-v2-clean & $79.2$ & $68.2$ & $85.5$ & $69.6$ && $57.5$ & $42.9$ & $67.2$ & $44.5$ \\
RRT (ours) & 500 & 200  & R50-v2-clean & $\num[math-rm=\mathbf]{79.5}$ & $\num[math-rm=\mathbf]{68.6}$ & $\num[math-rm=\mathbf]{87.8}$ & $\num[math-rm=\mathbf]{71.5}$ && $\num[math-rm=\mathbf]{62.5}$ & $\num[math-rm=\mathbf]{46.3}$ & $\num[math-rm=\mathbf]{77.1}$ & $\num[math-rm=\mathbf]{52.3}$ \\
\bottomrule
\end{tabular}
}
}
\vspace{-0.5em}
\caption{Comparison to geometric verification on Revisited Oxford/Paris~\cite{revisited}. 
The mAP scores on the Medium (+$\mathcal{R}$1M) and Hard (+$\mathcal{R}$1M) setups are reported.
Results marked by $\ast$ are evaluated by us using the public models provided by~\cite{delg2020}.
\vspace{-0.5em}
}
\label{tab:rt_vs_gv}
\end{table*}

\begin{table}[t]
\scalebox{0.95} {
    \centering
    \begin{tabular}{l l l c c}
    \toprule
    \multirow{2}{*}{Method} & \# local &Desc. & \multicolumn{2}{c}{\small Retrieval}\\
    &desc.&version &Public& Private\\
    \midrule
    DELG global  &0 &R50-v1& $18.3$ & $20.4$\\
    GV &1000&R50-v1& $20.4$ & $22.3$ \\
    RRT (ours)  &500&R50-v1& $\mathbf{21.5}$ & $\mathbf{23.1}$ \\
    \midrule
    DELG global  &0&R50-v2-clean& $22.2$ & $24.2$ \\
    GV &1000&R50-v2-clean& $-$ & $24.3$ \\
    RRT (ours)  &500&R50-v2-clean& $\mathbf{24.6}$ & $\mathbf{27.0}$ \\
    \bottomrule
    \end{tabular}
}
\vspace{-0.05in}
\caption{Comparison to geometric verification on the GLDv2 retrieval task~\cite{revisited}. The mAP@100 scores on the public and private test sets are reported.
\vspace{-0.15in}
}
\label{tab:gldv2}
\end{table}

\textbf{Experiments on SOP}:
We perform experiments on SOP~\cite{sop2016} using a single image scale, following the protocol for metric learning~\cite{XBM2020}.
During training, each image is randomly cropped to $224 \times 224$, followed by a random flip.
During testing,  each image is first resized to of $256 \times 256$ then cropped at the center to $224 \times 224$.
We use ResNet50 and extract features from the last convolutional layer, which leads to 49 ($7 \times 7$) local descriptors for each image.
The global descriptor is obtained by spatially averaging the local responses.
Both the global and local descriptors are linearly projected a dimension of 128. 
The RRT architecture and most of the training details remain the same as in the DELG experiments.
Here we only describe the main differences.
The global model is trained with a contrastive loss, as in~\cite{XBM2020}. 
Different from~\cite{XBM2020}, we do not rely on a cross batch memory but simply use a batch size of 800.
As all the local features are used, we do not incorporate the global descriptor term ($f_g(\mathbf{x}_g), \bar{f}_g(\mathbf{\bar{x}}_g)$) in Eq.~\ref{equ:input}.
We also drop the scale embedding ($\psi$) as only one image scale is used.
The global model is trained using SGD with Nesterov momentum for 100 epochs, using a learning rate of 0.001, a weight decay of 0.0004 and a momentum of 0.9. 
The learning rate drops by a factor of 10 after 60 and 80 epochs.
We train an RRT model on top of the pretrained global model, either freezing or finetuning the CNN backbone.
Both models are trained with AdamW~\cite{adamw2019} for 100 epochs, using a learning rate of 0.0001. 
The learning rate drops by a factor of 10 after 60 and 80 epochs. 
We implement RRTs in PyTorch~\cite{pytorch2019}. 

\textbf{Position embedding}:
For the experiments on DELG, where the keypoints are sparsely sampled, we observe no benefit in applying the position embedding and do not use the $\varphi$ term in Eq.~\ref{equ:embed}. For the experiments on SOP, we find the position embedding is helpful, we posit it is because all the positions are used in this experiment. 

\textbf{Latency and memory}
For each query, when using an NVIDIA P100 GPU, RRT reranks the top-100 retrieved images in a single forward-pass, which takes 0.36/0.013 seconds on average in the DELG~\cite{delg2020}/SOP~\cite{sop2016} experiments.
In the DELG experiments, we use the same global descriptor but only half (500 out of 1000) of the local descriptors for each image.
In other words, the memory footprint is approximately half of that in DELG~\cite{delg2020}. 
We agree that this is still a high cost for large-scale systems. 
In the future, we'd like to explore techniques that can potentially reduce the memory footprint, e.g. quantization.

\section{Results}
Here, we demonstrate the effectiveness of the Reranking Transformers (RRTs) across different settings, benchmarks and use cases.

\begin{table*}[t]
\setlength{\tabcolsep}{5pt}
\scalebox{0.9}{
\centering
{
\begin{tabular}{l l l c c c c c c c c c}
\toprule
\multirow{2}{*}{Method} & \# Reranked  & Desc. & \multicolumn{4}{c}{ Medium} && \multicolumn{4}{c}{ Hard} \\
& images &  version & \multicolumn{1}{c}{\large \vphantom{M} \normalsize $\mathcal{R}$Oxf } & \multicolumn{1}{c}{\normalsize +$\mathcal{R}$1M} & \multicolumn{1}{c}{\normalsize $\mathcal{R}$Par} & \multicolumn{1}{c}{\normalsize +$\mathcal{R}$1M} && \multicolumn{1}{c}{\normalsize $\mathcal{R}$Oxf} & \multicolumn{1}{c}{\normalsize +$\mathcal{R}$1M} & \multicolumn{1}{c}{\normalsize $\mathcal{R}$Par} & \multicolumn{1}{c}{\normalsize +$\mathcal{R}$1M} \\
\midrule
DELG global & & R50-v1 & $69.7$ & $55.0$ & $81.6$ & $59.7$ && $45.1$ &$27.8$ & $63.4$ & $34.1$ \\
\midrule
$\alpha$QE &  & R50-v1 & $72.9$ &$60.7$ & \underline{$83.4$} & \underline{$63.7$} && $49.4$ & $33.6$ & $66.1$ & \underline{$38.1$} \\
RRT (ours) & 100 & R50-v1 & $75.5$ &$61.2$ & $82.7$ & $60.7$ && $56.4$ & $37.0$ & $68.6$ & $37.5$ \\
RRT (ours) & 200 & R50-v1 & $77.9$ &$63.5$ & $84.4$ & $62.1$ && $58.8$ & $39.5$ & $71.6$ & $39.5$ \\ 
RRT (ours) & 400 & R50-v1 & $\num[math-rm=\mathbf]{79.2}$ & $\num[math-rm=\mathbf]{66.2}$ & $\num[math-rm=\mathbf]{86.3}$ & $\num[math-rm=\mathbf]{64.0}$ && $\num[math-rm=\mathbf]{60.5}$ & $\num[math-rm=\mathbf]{42.6}$ & $\num[math-rm=\mathbf]{74.1}$ & $\num[math-rm=\mathbf]{41.6}$ \\
\midrule
$\alpha$QE &  & R50-v1 & $72.9$ &$60.7$ & $83.4$ & $63.7$ && $49.4$ & $33.6$ & $66.1$ & $38.1$ \\
$\alpha$QE + RRT (ours) & 200 & R50-v1 & \num[math-rm=\mathbf]{78.7} & \num[math-rm=\mathbf]{66.2} & \num[math-rm=\mathbf]{85.6} & \num[math-rm=\mathbf]{65.4} && \num[math-rm=\mathbf]{59.8} & \num[math-rm=\mathbf]{42.1} & \num[math-rm=\mathbf]{72.8} & \num[math-rm=\mathbf]{43.1} \\
\midrule
DELG global & & R50-v2-clean & $73.6$ & $60.6$ & $85.7$ & $68.6$ && $51.0$ & $32.7$ & $71.5$ & $44.4$ \\
\midrule
$\alpha$QE &  & R50-v2-clean & $76.6$ &$66.4$ & $86.7$ & \underline{$72.8$} && $54.6$ & $39.5$ & $73.2$ & \underline{$51.2$} \\
RRT (ours) & 100 & R50-v2-clean & $78.1$ & $67.0$ & $86.7$ & $69.8$ && $60.2$ & $44.1$ & $75.1$ & $49.4$   \\
RRT (ours) & 200 & R50-v2-clean & $79.5$ & $68.6$ & $87.8$ & $71.5$ && $62.5$ & $46.3$ & $77.1$ & $52.3$ \\
RRT (ours) & 400 & R50-v2-clean & $\num[math-rm=\mathbf]{80.5}$ & $\num[math-rm=\mathbf]{70.6}$ & $\num[math-rm=\mathbf]{89.1}$ & $\num[math-rm=\mathbf]{73.8}$ && $\num[math-rm=\mathbf]{64.2}$ & $\num[math-rm=\mathbf]{49.5}$ & $\num[math-rm=\mathbf]{78.1}$ & $\num[math-rm=\mathbf]{55.6}$ \\
\midrule
$\alpha$QE &  & R50-v2-clean & $76.6$ &$66.4$ & $86.7$ & $72.8$ && $54.6$ & $39.5$ & $73.2$ & $51.2$ \\
$\alpha$QE + RRT (ours) & 200 & R50-v2-clean & \num[math-rm=\mathbf]{80.4} & \num[math-rm=\mathbf]{71.7} & \num[math-rm=\mathbf]{88.5} & \num[math-rm=\mathbf]{74.8} && \num[math-rm=\mathbf]{64.0} & \num[math-rm=\mathbf]{50.9} & \num[math-rm=\mathbf]{77.7} & \num[math-rm=\mathbf]{57.1}\\
\bottomrule
\end{tabular}
}
}
\vspace{-0.5em}
\caption{Comparison to $\alpha$QE~\cite{finetune2016} on Revisited Oxford/Paris~\cite{revisited}. The mAP scores on the Medium (+$\mathcal{R}$1M) and Hard (+$\mathcal{R}$1M) setups are reported. We underline the scores of $\alpha$QE that RRT cannot match by just reranking the top-100 neighbors. RRT consistently outperforms $\alpha$QE when reranking the top-400 neighbors for each query. Moreover, combining $\alpha$QE with RRT significantly outperforms using $\alpha$QE only, showing that RRT and $\alpha$QE are complementary to each other.
\vspace{-1.5em}
}
\label{tab:rt_vs_aqe}
\end{table*}
\begin{table}[t]
\scalebox{0.88} {
\setlength{\tabcolsep}{2pt}
\centering
{
\begin{tabular}{l l l c c c c c}
\toprule
\multirow{2}{*}{Method} & \# local & Desc. & \multicolumn{2}{c}{ Medium} && \multicolumn{2}{c}{ Hard} \\
& desc. &version & \multicolumn{1}{c}{\normalsize \vphantom{M} \normalsize $\mathcal{R}$Oxf } & \multicolumn{1}{c}{\normalsize $\mathcal{R}$Par} && \multicolumn{1}{c}{\normalsize $\mathcal{R}$Oxf} &  \multicolumn{1}{c}{\normalsize $\mathcal{R}$Par}\\
\midrule
DELG global & 0 & R50-v1 & $69.7$ & $81.6$ & & $45.1$ & $63.4$\\
ASMK global & 1000 & R50-v1 & $71.2$ & $80.8$ && $47.1$ & $61.6$\\
ASMK rerank & 1000 & R50-v1 & $71.3$ & $82.6$ && $47.5$ & $66.2$\\
RRT (ours) & 500 & R50-v1 & $\num[math-rm=\mathbf]{75.5}$ & $\num[math-rm=\mathbf]{82.7}$ && $\num[math-rm=\mathbf]{56.4}$ & $\num[math-rm=\mathbf]{68.6}$\\
\midrule
DELG global & 0 & R50-v2-clean & $73.6$ & $85.7$ && $51.0$ & $71.5$\\
ASMK global & 1000 & R50-v2-clean & $70.4$ & $80.9$ && $45.8$ & $62.0$\\
ASMK rerank & 1000 & R50-v2-clean & $73.1$ & $86.3$ && $49.3$ & $71.9$\\
RRT (ours) & 500 & R50-v2-clean & $\num[math-rm=\mathbf]{78.1}$ & $\num[math-rm=\mathbf]{86.7}$ && $\num[math-rm=\mathbf]{60.2}$ & $\num[math-rm=\mathbf]{75.1}$\\
\bottomrule
\end{tabular}
}
}
\vspace{-0.5em}
\caption{Comparison to Aggregated Selective Match Kernel (ASMK) on Revisited Oxford/Paris~\cite{revisited}. 
The mAP scores on the Medium and Hard setups are reported.
\vspace{-1.5em}
}
\label{tab:rt_vs_asmk}
\end{table}

\subsection{Comparison with Geometric Verification}
We consider geometry verification (GV) as the main baseline.
We compare GV and RRT using the same pretrained DELG~\cite{delg2020} descriptors.
Following the protocol in~\cite{delg2020}, given a query, we use its global descriptor to retrieve a set of top-ranked images. 
The top-100 neighbors are reranked by GV and RRT.
We present results on two sets of descriptors: DELG pretrained on GLD v1~\cite{delf2017} and v2-clean~\cite{gldv2}.

On $\mathcal{R}$Oxf and $\mathcal{R}$Par, both GV and RRT outperform global-only retrieval, as shown in Table~\ref{tab:rt_vs_gv}. 
RRT shows further advantages over GV, with much fewer local descriptors. 
On $\mathcal{R}$Oxf (+$\mathcal{R}$1M), RRT performs on par with GV on the Medium setup and consistently better on the Hard setup.
On $\mathcal{R}$Par (+$\mathcal{R}$1M), RRT consistently outperforms GV.
The largest performance gap is achieved on the Hard setup.
RRT obtains 2.2 (3.7) absolute improvements over GV on $\mathcal{R}$Oxf ($\mathcal{R}$Par), when using the ``v1'' descriptors.
We posit that, while GV is effective for sufficiently similar images, it has difficulty handling challenging cases, e.g. large variations in viewpoint.
To verify this, we reranked more images (e.g.~top-200), resulting in a larger performance gap.
RRT obtains 3.4 (8.4) absolute improvement over GV on $\mathcal{R}$Oxf ($\mathcal{R}$Par), when using the ``v1'' descriptors.

We present results on the GLDv2 retrieval task~\cite{gldv2} in Table~\ref{tab:gldv2}.
Following~\cite{delg2020}, we report the mAP@100 scores on the public and private test sets.
Compared to $\mathcal{R}$Oxf and $\mathcal{R}$Par, the improvement of applying reranking on GLDv2 becomes smaller.
On the other hand, RRT performs consistently better than global-only and GV.
When using the ``v2-clean'' descriptors, the absolute improvements of RRT over global-only (GV) on the private set are 2.8 (2.7).


\begin{table*}[t]
\scalebox{0.86} {
    \setlength{\tabcolsep}{2.5pt}
    \centering
    {\begin{tabular}{l l l c c c c c c c c c c c c }
    \toprule
     \multirow{3}{*}{Method} & \multirow{3}{*}{Training set} & \multirow{3}{*}{Net} & \multirow{3}{*}{\begin{tabular}{@{}c@{}}\# local \\ desc.\end{tabular}} & \multicolumn{4}{c}{ Medium} && \multicolumn{4}{c}{ Hard} \\
    \cmidrule[0.5pt]{5-8} \cmidrule[0.5pt]{10-13} 
     &&&& \multicolumn{1}{c}{\large \vphantom{M} \normalsize $\mathcal{R}$Oxf } & \multicolumn{1}{c}{\normalsize +$\mathcal{R}$1M} & \multicolumn{1}{c}{\normalsize $\mathcal{R}$Par} & \multicolumn{1}{c}{\normalsize +$\mathcal{R}$1M} && \multicolumn{1}{c}{\normalsize $\mathcal{R}$Oxf} & \multicolumn{1}{c}{\normalsize +$\mathcal{R}$1M} & \multicolumn{1}{c}{\normalsize $\mathcal{R}$Par} & \multicolumn{1}{c}{\normalsize +$\mathcal{R}$1M} \\
    \midrule
    { \textit{(A) Global features}} \\
    R-MAC \cite{DIR2017} & Landmarks & R101 & 0  &   \num{60.9}  & \num{39.3} & \num{78.9} & \num{54.8} && \num{32.4} & \num{12.5} & \num{59.4} & \num{28.0} \\ 
    GeM \cite{finetune2016} & SfM-120k & R101 & 0 & \num{64.7}  & \num{45.2} & \num{77.2}  & \num{52.3} && \num{38.5} & \num{19.9} & \num{56.3} & \num{24.7} \\
    GeM-AP \cite{listwise2019} & SfM-120k & R101 & 0  & \num{67.5} & \num{47.5} & \num{80.1} & \num{52.5} && \num{42.8} & \num{23.2} & \num{60.5} & \num{25.1} \\
    DELG~\cite{delg2020} & GLDv1 & R50 & 0  & \num{69.7} & \num{55.0} & \num{81.6} & \num{59.7} && \num{45.1} & \num{27.8} & \num{63.4} & \num{34.1} \\
    \midrule
    { \textit{(B) Local feature aggregation}} \\ 
    DELF-ASMK\cite{rasmk2019} & Landmarks & R50 & 1000 & \num{67.8}  & \num{53.8} & \num{76.9} & \num{57.3} && \num{43.1} & \num{31.2} & \num{55.4} & \num{26.4} \\
    HOW-ASMK\cite{how2020} & SfM-120k & R50 & 1000 & \num{78.3}  & \num{63.6} & \num{80.1} & \num{58.4} && \num{55.8} & \num{36.8} & \num{60.1} & \num{30.7} \\ 
    HOW-ASMK\cite{how2020} & SfM-120k & R50 & 2000 & \num[math-rm=\mathbf]{79.4}  & \num{65.8} & \num{81.6} & \num{61.8} && \num{56.9} & \num{38.9} & \num{62.4} & \num{33.7} \\ 
    \midrule
    { \textit{(C) Global features + Re-ranking} }\\
    
    GeM$\uparrow$+DSM \cite{dsm2019} & SfM-120k& R101 & 1000 & \num{65.3} & \num{47.6} & \num{77.4} & \num{52.8} &&  \num{39.2} & \num{23.2} & \num{56.2} & \num{25.0} \\
    
    DELG~\cite{delg2020} + GV & GLDv1 & R50 & 1000 & $75.1$ &$61.1$ & $82.3$ & $60.5$ && $54.2$ & $36.8$ & $64.9$ & $34.8$ \\
    
    DELG~\cite{delg2020} + RRT (ours) & GLDv1\&v2-clean & R50 & 500 & $\num{75.5}$ &$\num{61.2}$ & $\num{82.7}$ & $\num{60.7}$ && $\num{56.4}$ & $\num{37.0}$ & $\num{68.6}$ & $\num{37.5}$ \\ 
    
    DELG~\cite{delg2020} + GV & GLDv2-clean & R50 & 1000 & $\num{78.3}$ & $\num[math-rm=\mathbf]{67.2}$ & $85.7$ & $69.6$ && $57.9$ & $43.6$ & $71.0$ & $45.7$ \\
    
    DELG~\cite{delg2020} + RRT (ours) & GLDv2-clean & R50 & 500 & $78.1$ & $67.0$ & $\num[math-rm=\mathbf]{86.7}$ & $\num[math-rm=\mathbf]{69.8}$ && $\num[math-rm=\mathbf]{60.2}$ & $\num[math-rm=\mathbf]{44.1}$ & $\num[math-rm=\mathbf]{75.1}$ & $\num[math-rm=\mathbf]{49.4}$ \\ 
    
    \bottomrule
    \end{tabular}
    }
}
\caption{Comparison to the state-of-the-art on Revisited Oxford/Paris~\cite{revisited}. The mAP scores on the Medium and Hard setups are reported.
\vspace{-0.1in}
}
\label{tab:sota}
\end{table*}

\begin{table}[t]
\setlength{\tabcolsep}{1.5pt}
    \begin{tabular}{lccccc}
    \toprule
    \multirow{2}{*}{Method} & Desc. & \multicolumn{4}{c}{SOP} \\ 
    & dim. & ${R@1}$ & ${R@10}$ & ${R@100}$ & ${R@1k}$ \\
    \midrule
    { \textit{Global-only retrieval}} \\
    Margin~\cite{margin2017,roth2020revisiting}  & 128 & 76.1 & 88.4 & 95.1 & 98.3\\
    FastAP~\cite{fastap2019}  & 128 & 73.8 & 88.0 & 94.9 & 98.3\\
    XBM~\cite{XBM2020}  & 128 & 80.6 & 91.6 & 96.2 & 98.7 \\
    CE~\cite{CE2020} & 2048 & 81.1 & 91.7 & 96.3 & 98.8 \\
    \midrule
    CO & 128 & 80.7 & 91.9 & \num[math-rm=\mathbf]{96.6} & \num[math-rm=\mathbf]{99.0}\\
    CO + RRT (frozen) & 128 & 81.8 & 92.4 & \num[math-rm=\mathbf]{96.6} &\num[math-rm=\mathbf]{99.0}  \\
    CO + RRT (finetuned)& 128 & \num[math-rm=\mathbf]{84.5} & \num[math-rm=\mathbf]{93.2} & \num[math-rm=\mathbf]{96.6} &\num[math-rm=\mathbf]{99.0}\\
    \bottomrule
    
    \end{tabular}
\vspace{-0.5em}
\caption{Results on jointly optimizing the feature extractor and RRT. The R@K (K =1, 10, 100, 1000) scores on the SOP test set~\cite{sop2016} are reported.
\vspace{-0.14in}
}
\label{tab:sop}
\end{table}

\subsection{Comparison with Query Expansion}

Query expansion (QE)~\cite{qe2007, qe2011, qe2014} is another popular reranking technique for image retrieval. 
Different from GV and RRT, QE aggregates the query image and a number of top-ranked neighbors into a new query.
This new query is used to rerank all the gallery images rather than the nearest ones as in GV and RRT.
We compare RRT with one of the most widely used query expansion methods: $\alpha$-weighted query expansion ($\alpha$QE) proposed in~\cite{finetune2016}.
We use the public implementation of $\alpha$QE
released by~\cite{listwise2019}.
$\alpha$QE has two hyper-parameters: (1) nQE, the number of top-ranked neighbors to aggregate; (2) $\alpha$, the exponential weight.
In~\cite{listwise2019}, they are set as $(n\textsc{QE}, \alpha) = (10, 2.0)$.
Our experiment shows that these values do not work out of the box for the DELG descriptors.
We tune these parameters on $\mathcal{R}$Oxf over the ranges: $n\textsc{QE} \in [2,15], \alpha \in [0.1,3.0]$, and eventually set them as $(n\textsc{QE}, \alpha) = (2, 0.3)$.

Table~\ref{tab:rt_vs_aqe} shows the results on $\mathcal{R}$Oxf and $\mathcal{R}$Par.
When reranking the top 100 neighbors, the performance of RRT is superior to $\alpha$QE on five of the eight settings, except for $\mathcal{R}$Par+Medium, $\mathcal{R}$Par+$\mathcal{R}$1M+Medium, $\mathcal{R}$Par+$\mathcal{R}$1M+Hard (underlined numbers).
We believe it is because $\alpha$QE reranks all the gallery images while RRT reranks only 100 neighbors and keeps the ranks of all the other images unchanged.
By reranking more neighbors, e.g. 200, 400, we observe that the performance of RRT progressively improves and eventually surpasses $\alpha$QE by significant margins across all settings.
On the Hard setup with the ``v1'' descriptors, the absolute gains of RRT over $\alpha$QE on ($\mathcal{R}$Oxf, $\mathcal{R}$Oxf+$\mathcal{R}$1M, $\mathcal{R}$Par, $\mathcal{R}$Par+$\mathcal{R}$1M) are (11.1, 9.0, 8.0, 3.5).

We also perform experiments on combining $\alpha$QE and RRT by reranking the top neighbors produced by $\alpha$QE.
As shown in Table~\ref{tab:rt_vs_aqe}, reranking the top-200 images obtained from $\alpha$QE considerably improves over using $\alpha$QE only, with improvements of (10.4, 8.5, 6.7, 5.0) on the Hard setup of ($\mathcal{R}$Oxf, $\mathcal{R}$Oxf+$\mathcal{R}$1M, $\mathcal{R}$Par, $\mathcal{R}$Par+$\mathcal{R}$1M) for the ``v1'' descriptors.
We consider query expansion and RRT are thus complementary.

\subsection{Comparison with Aggregated Selective Match Kernel (ASMK)}

Aggregated Selective Match Kernel (ASMK)~\cite{asmk2016} also leverages local descriptors for image retrieval. 
The key idea is to create a large visual codebook (i.e. filter banks) by clustering the local descriptors.
This visual codebook is used to encode the query and gallery images into global descriptors. 
The clustering and encoding procedures are typically performed offline as they're relatively time-consuming.
Previously, ASMK was mainly considered as a global retrieval technique. 
In this paper, we treat ASMK as both a global retrieval baseline and a reranking baseline.
We use the public implementation of ASMK released by~\cite{rasmk2019}.
Following the common practice proposed in~\cite{rasmk2019}, we train a codebook of 65,536 visual words on $\mathcal{R}$Oxf for retrieval experiments on $\mathcal{R}$Par, and vice-versa. 
We conduct two experiments: a) ASMK global: using ASMK for global retrieval, as in all the previous literature~\cite{asmk2016, rasmk2019, how2020}; b) ASMK rerank: using ASMK for image reranking, e.g. reranking the top-100 images from DELG global retrieval. 

\begin{figure*}[t]
  \centering
  \includegraphics[width=0.76\textwidth]{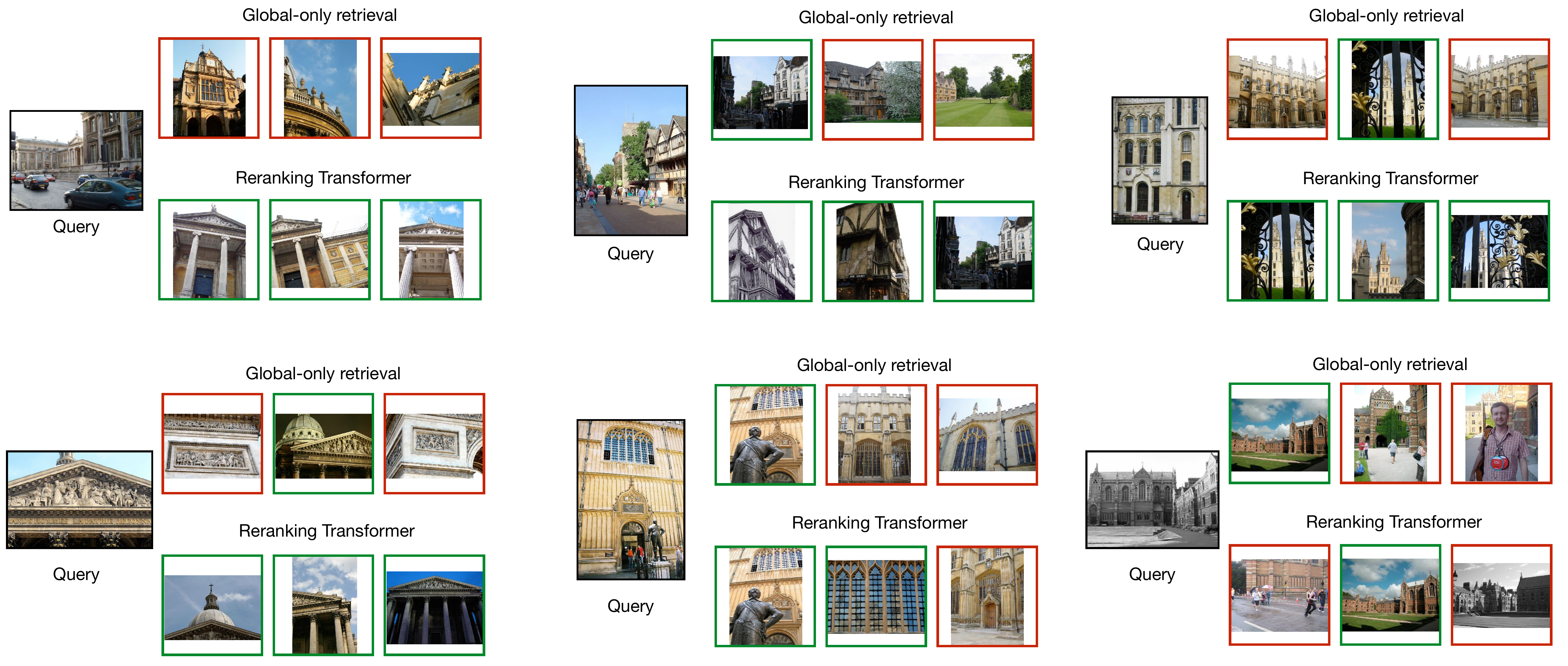}
  \vspace{-0.06in}
  \caption{Qualitative examples from Revisited Oxford/Paris~\cite{revisited}. For each query, the top-3 neighbors ranked by the global retrieval and reranked by RRT are presented. Correct/incorrect neighbors are marked with {\color{green}green}/{\color{red}red} borders.
  \vspace{-0.12in}
  }
  \label{fig:qual}
\end{figure*}

We present the results on $\mathcal{R}$Oxf and $\mathcal{R}$Par in Table~\ref{tab:rt_vs_asmk}. 
ASMK, when used as a global retrieval approach, shows comparable or inferior performance to DELG global.
When used as a reranking approach, ASMK further improves over DELG global, showing that they are complementary.
The proposed method consistently outperforms ASMK global/rerank in all settings.
We posit that compared with the hand-crafted kernel matching paradigm, RRTs learn a more holistic region-wise similarity between the images.

\subsection{Feature Learning \& RRT: Joint Optimization}
To further demonstrate the possibility of jointly optimizing the feature representations and RRTs, 
we perform experiments on the Stanford Online Products (SOP) dataset~\cite{sop2016}.
We study three models: 
(1) \textit{CO}: A global retrieval model trained with a contrastive loss~\cite{XBM2020}, using the metric learning protocol,~i.e.~the global descriptor has a dimension of 128~\cite{roth2020revisiting,fastap2019,divide2019, MIC2019, XBM2020}; (2) \textit{CO + RRT (frozen)}: an RRT model trained on top of \textit{CO}. 
The pretrained \textit{CO} remains frozen and an extra linear layer is used to reduce the dimension of the local descriptors to 128; 
(3) \textit{CO + RRT (finetune)}: a model with the same architecture as \textit{CO + RRT (frozen)} but the backbone is also finetuned. 
It is also initialized by \textit{CO + RRT (frozen)}. 
During testing, we perform global retrieval using the global descriptor from \textit{CO}.
The top-100 neighbors for each query are reranked by either \textit{CO + RRT (frozen)} or \textit{CO + RRT (finetune)}.
We present our results along with the results of the most recent metric learning approaches~\cite{roth2020revisiting, fastap2019, XBM2020, CE2020} to provide an overview in the context of the state-of-the-art on SOP.

As shown in Table~\ref{tab:sop}, the global \textit{CO} model, which is trained with a contrastive loss using a relatively large batch size, performs surprisingly well. 
It achieves the same level of accuracy as well-established works on metric learning.
This aligns with the recent research on self-supervised learning~\cite{simclr2020, moco2020} showing that contrastive loss is very effective for feature learning.
\textit{CO + RRT (frozen)} further improves the performance, demonstrating the effectiveness of reranking. 
Note that, as only the top-100 images are reranked, the R@100 and R@1k scores remain unchanged.
\textit{CO + RRT (finetuned)} achieves the best reranking performance, with an absolute improvement of 3.8 over the global-only retrieval on R@1. 
We believe it is because jointly optimizing the backbone and our model leads to better local features that are tailored to the reranking tasks.


\subsection{Comparison with the State-of-the-Art}

In Table~\ref{tab:sota}, we compare the proposed method with the state-of-the-art on the $\mathcal{R}$Oxf (+$\mathcal{R}$1M) and $\mathcal{R}$Par (+$\mathcal{R}$1M) benchmarks. 
We include the most recent instance recognition/retrieval models in three different groups: (A) Retrieval by global features only; (B) Retrieval by local feature aggregation; (C) Retrieval by combining global features with reranking.
While our method performs favorably on most of the settings (except for $\mathcal{R}$Oxf, $\mathcal{R}$Oxf+$\mathcal{R}$1M), these results include comparisons to methods that differ on the training data, the CNN backbones, and the number of local features, etc. For context we provide as much information about each method regarding these differences.

In Fig.~\ref{fig:qual}, we present qualitative examples on image retrieval when using only global features and when using our full reranking approach. While global-only retrieval can return highly similar images in general, reranking by global/local descriptors captures a more fine-grained matching between images, leading to better recognition accuracy.

Finally, Table~\ref{tab:rrt_vs_glue_revisited} shows a comparison with SuperGlue~\cite{superglue2020}.  
Similar as in geometry verification, the number of inlier correspondences predicted by SuperGlue is used as the similarity score. 
As SuperGlue is not designed for global retrieval, we use the pretrained DELG v1/v2-clean descriptors for global retrieval, so that SuperGlue and our method are evaluated on the same initial ranking lists. 
We use the SuperGlue pretrained on MegaDepth~\cite{megadepth2018}, which contains 130K images with dense annotations. 
Note that both SuperGlue and our work are trained on datasets (MegaDepth vs a subset of GLDv2-clean) that have different data distributions than the test sets (Revisited Oxford/Paris). 
SuperGlue leverages SuperPoint~\cite{superpoint2018} as the backbone, and uses different feature dimensions and keypoint numbers as our work. 
As shown in Table~\ref{tab:rrt_vs_glue_revisited},
reranking by both SuperGlue and our work significantly improve the performance over the global-only retrieval. RRT gains larger improvements on most of the settings, especially for challenge cases. We provide more experiments on SuperGlue in the supplemental material.

\begin{table}[t]
\scalebox{0.71}{
\setlength{\tabcolsep}{3pt}
\centering
{
\begin{tabular}{l l l l c c c c c}
\toprule
\multirow{2}{*}{Method} & Desc. & \# local & Desc.  & \multicolumn{2}{c}{ Medium} && \multicolumn{2}{c}{ Hard} \\
 & version & desc. & dim. & \multicolumn{1}{c}{\normalsize \vphantom{M} \normalsize $\mathcal{R}$Oxf } & \multicolumn{1}{c}{\normalsize $\mathcal{R}$Par} && \multicolumn{1}{c}{\normalsize $\mathcal{R}$Oxf} & \multicolumn{1}{c}{\normalsize $\mathcal{R}$Par}\\
\midrule
DELG global & R50-v1 & 0 & - & $69.7$ & $81.6$ && $45.1$ & $63.4$ \\
\midrule
SuperGlue~\cite{superglue2020} & SuperPoint~\cite{superpoint2018} & 500 & 256 & $72.6$ & ${81.9}$ && $49.7$ & ${62.4}$ \\
SuperGlue~\cite{superglue2020} & SuperPoint~\cite{superpoint2018} & 1024 & 256 & $74.4$ & ${82.3}$ && $55.2$ & ${64.6}$ \\
RRT (ours) & R50-v1 & 500 & 128 & $\num[math-rm=\mathbf]{75.5}$ & $\num[math-rm=\mathbf]{82.7}$ && $\num[math-rm=\mathbf]{56.4}$ & $\num[math-rm=\mathbf]{68.6}$ \\
\midrule
DELG global & R50-v2-clean & 0 &- & $73.6$ & $85.7$ && $51.0$ & $71.5$ \\
\midrule
SuperGlue~\cite{superglue2020} & SuperPoint~\cite{superpoint2018} & 500 & 256 & $76.2$ & ${85.9}$ && $54.6$ & ${68.2}$  \\
SuperGlue~\cite{superglue2020} & SuperPoint~\cite{superpoint2018} & 1024 & 256 & $\num[math-rm=\mathbf]{78.3}$ & ${86.2}$ && $60.0$ & ${70.4}$  \\
RRT (ours)  & R50-v2-clean & 500 & 128 & ${78.1}$ & $\num[math-rm=\mathbf]{86.7}$ && $\num[math-rm=\mathbf]{60.2}$ & $\num[math-rm=\mathbf]{75.1}$ \\
\bottomrule
\end{tabular}
}
}
\vspace{-0.05in}
\caption{Comparison to the pretrained SuperGlue model~\cite{superglue2020} on Revisited Oxford/Paris~\cite{revisited}. 
The SuperGlue model is pretrained on MegaDepth~\cite{megadepth2018} with SuperPoint~\cite{superpoint2018} as the backbone.
The mAP scores on the Medium and Hard setups are reported.
\vspace{-0.15in}
}
\label{tab:rrt_vs_glue_revisited}
\end{table}

\section{Conclusion}
We introduce \textit{Reranking Transformers} (RRTs) for instance image retrieval.
We show that RRTs outperform prior reranking approaches across a variety of settings.
Compared to geometric verification~\cite{spatial2007} and other local feature based methods~\cite{asmk2016}, RRTs use fewer descriptors and can be parallelized such that reranking requires a single forward pass.
We also demonstrate that, unlike previous reranking approaches, RRTs can be optimized jointly with the feature extractor, leading to further gains. 


\vspace{0.05in}
\noindent{\bf Acknowledgements:}~This project was funded by a gift from eBay Inc to the University of Virginia. We also thank Robinson Piramuthu for his encouragement in starting this collaboration, Dmytro Mishkin for suggesting comparisons with SuperGlue and anonymous reviewers for their feedback. V.O. was also supported by a Facebook Research Award, a gift from Leidos Inc, and NSF Award IIS-2045773.

\clearpage
\begin{figure}[ht!]
  \begin{flushleft}
  \Large\textbf{Supplementary Material}
  \end{flushleft}
\end{figure}
This document is organized as follows.
In Sec.~\ref{sec:baseline}, we discuss why we consider geometry verification, query expansion, and aggregated selective match kernel as the baseline methods. 
In Sec.~\ref{sec:ablation}, we provide an ablation study on using different numbers of local descriptors in geometry verification (GV)~\cite{spatial2007} and RRT. 
In Sec.~\ref{sec:superpoint}, we perform experiments using SuperPoint~\cite{superpoint2018} as the feature extractor for RRT, and compare with SuperGlue~\cite{superglue2020} on Stanford Online Products~\cite{sop2016}.
In Sec.~\ref{sec:r101}, we perform experiments using ResNet101~\cite{resnet2016} as the CNN backbone.
In Sec.~\ref{sec:correspondences}, we visualize the keypoint correspondences learned by RRT.
In Sec.~\ref{sec:limitation}, we discuss the limitation of the proposed method.
Finally, in Sec.~\ref{sec:examples}, we present more qualitative examples.

The names of the training images sampled from GLDv2, as discussed in Section 4.1 of the main paper, are in a separate document.

\section{Appropriate baselines}
\label{sec:baseline}

We consider geometry verification~\cite{spatial2007} and $\alpha$QE~\cite{qe2007} as the main baselines as they share the same spirit with our method: they make better use of the test-time information.
When comparing the query and target images, geometry verification attends to \textit{different} sub-regions of the query image when the target image is \textit{different}, and vice versa, which is very similar to the proposed Reranking Transformers (RRTs).
$\alpha$QE also leverages test-time knowledge, but relies on analyzing the local affinity graph created during testing.
We believe incorporating test-time knowledge is the key motivation of image reranking. 
It also distinguishes our method from most of the previous approaches that focus on feature learning. Note that we use pretrained and fixed feature representation in most of our experiments. 

\begin{figure}[t]
  \centering
  \includegraphics[width=0.9\textwidth]{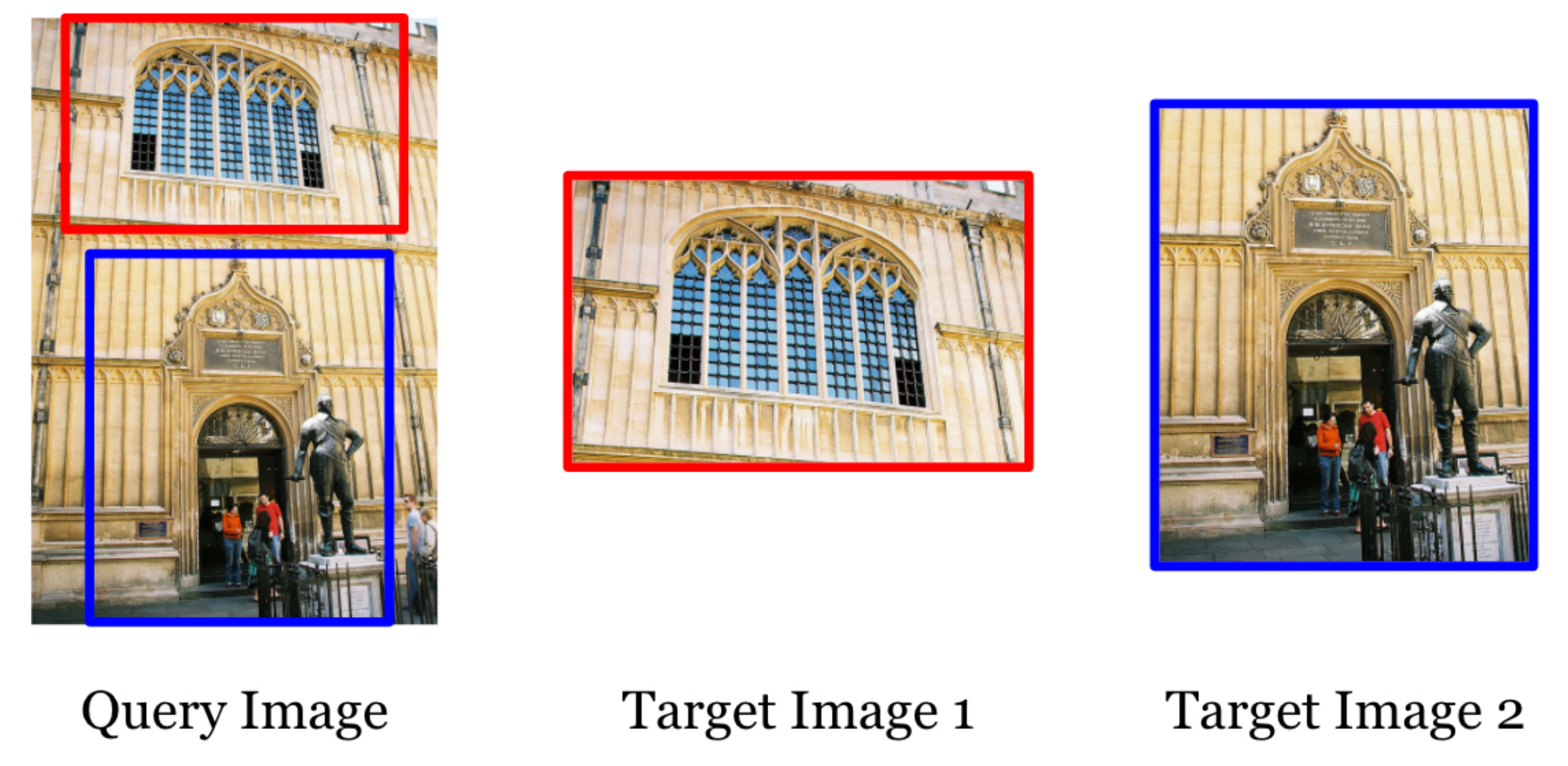}
  \caption{An example where the target images are some crops of the query. In this case the global descriptor + cosine similarity retrieval paradigm may not be ideal.}
  \label{fig:intuition}
\end{figure}

Fig.~\ref{fig:intuition} provides an intuitive example of the partial-matching cases.
In this example, the target images are some crops of the query. 
We believe the global descriptor + cosine similarity paradigm is not ideal for this case, as no matter how large is the global descriptor, it contains irrelevant information that hinders the cosine similarity measurement.


Aggregated Selective Match Kernel (ASMK)~\cite{asmk2016} was previously used as a global retrieval approach instead of an image reranking approach. 
Specifically, it proposes to create a set of new filters (i.e. visual codebook) by clustering. 
It then remaps/aggregates the local descriptors of each image into a global vector. 
We perform experiments on ASMK as it also relies on local descriptors.

\section{Ablation on the number of local descriptors}
\label{sec:ablation}
In the DELG model, for each image, a maximum of 1000 local descriptors are extracted for geometric verification. 
In our experiment, we observe that for most of the images, the number of local descriptors is close to 1000. 
For example, on the sampled GLDv2 training set, the query and gallery sets of Revisited Oxford ($\mathcal{R}$Oxf)~\cite{revisited}, DELG extracts 955/759/987 local descriptors per image on average. 

We perform an ablation experiment by setting the maximum number of local descriptors used for each image to different values. 
The DELG model~\cite{delg2020} used in this experiment is pretrained on the ``v2-clean'' split of Google Landmarks v2 (GLDv2)~\cite{gldv2}. 
For purposes of comparison, we include the results of geometry verification (GV) and the proposed method (RRT).
We report the mAP scores on Revisited Oxford ($\mathcal{R}$Oxf) in Table~\ref{tab:ablation}. 

Both GV and RRT benefit from using more local descriptors in general.
Nevertheless, the performance of RRT saturates at 500 local descriptors. 
As the local descriptors are extracted from seven image scales, we conjecture that in each image there are descriptors extracted from the same location, thus providing duplicate information. 
To verify this, we compute the number of \textit{distinct} local descriptors extracted from different grid locations. 
In particular, we assign each local descriptor $\mathbf{x}_{l, i}$ to a grid location $(gu, gv)$ by $(gu, gv) = (\lfloor u / 16\rfloor, \lfloor v / 16 \rfloor)$.
Here $(u, v)$ is the coordinate of $\mathbf{x}_{l, i}$ provided by the DELG model, \num{16} is the stride of the convolutional feature map where $\mathbf{x}_{l, i}$ is extracted from. 
We then group the descriptors sharing the same grid location as a \textit{distinct} descriptor.
We observe that, the number of \textit{distinct} local descriptors is significantly smaller than the number of all local descriptors per image.
For example, on the sampled GLDv2 training set, the query and gallery sets of Revisited Oxford ($\mathcal{R}$Oxf), the numbers of \textit{distinct} local descriptors per image are 585/465/655 on average.

When using the same number of local descriptors, RRT outperforms GV in four of the six experiments on the Medium setup, and consistently outperforms GV on the Hard setup. 

\begin{table}[t]
\centering
{
\begin{tabular}{l c c c c c}
\toprule
\# Local & \multicolumn{2}{c}{ Medium} && \multicolumn{2}{c}{ Hard}\\
Desc.  & \multicolumn{1}{c}{\normalsize GV} & \multicolumn{1}{c}{\normalsize RRT} && \multicolumn{1}{c}{\normalsize GV} & \multicolumn{1}{c}{\normalsize RRT}\\
\midrule
 200 & \num{72.1} &\num{76.7}&& \num{48.3}&\num{58.9}\\
 400 & \num{75.2} &\num{77.6}&& \num{53.8}&\num{58.6}\\
 500 & \num{75.7} &\num{78.1}&& \num{53.4}&\num{60.2}\\
 600 & \num{77.4} &\num{77.9}&& \num{55.9}&\num{59.6}\\
 800 & \num{77.9} &\num{76.9}&& \num{56.7}&\num{57.4}\\
1000 & \num{78.3} &\num{78.1}&& \num{57.9}&\num{60.4}\\
\bottomrule
\end{tabular}
}
\caption{Ablation on the number of local descriptors used per image. We compare the proposed Reranking Transformer (RRT) model to geometric verification (GV) on Revisited Oxford~\cite{revisited}. 
The mAP scores on the Medium  and Hard setups are reported.}
\label{tab:ablation}
\end{table}

\section{SuperPoint as the CNN backbone.}
\label{sec:superpoint}
\begin{table}[t]
\setlength{\tabcolsep}{1.5pt}
    \begin{tabular}{lcccc}
    \toprule
    Method  & ${R@1}$ & ${R@10}$ & ${R@100}$ \\
    \midrule
    Global-only & 32.8 & 45.4 & 60.5\\
    \midrule
    SuperGlue~\cite{superglue2020} & 45.5 & 54.6 & 60.5\\
    \midrule
    RRT (w pos, frozen) & 47.3 & 56.5 & 60.5\\
    RRT (w/o pos, frozen) & 50.2 & 57.9 & 60.5\\
    RRT (w/o pos, finetuned) & \num[math-rm=\mathbf]{51.9} & \num[math-rm=\mathbf]{59.0} & {60.5}\\
    \bottomrule
    \end{tabular}
\caption{Comparison to the pretrained SuperGlue model~\cite{superglue2020} on Stanford Online Products~\cite{sop2016}, using SuperPoint~\cite{superpoint2018} as the CNN backbone. The SuperGlue model is pretrained on ScanNet~\cite{scannet2017}. The R@K (K =1, 10, 100) scores on the SOP~\cite{sop2016} test set are reported. Note that as only the top-100 neighbors are reranked, the R@100 scores remain unchanged for all the models.}
\label{tab:rrt_vs_glue_sop}
\end{table}
In the main paper, we compare Reranking Transformer (RRT) with SuperGlue~\cite{superglue2020} on Revisited Oxford/Paris, but the feature extractors used for the two models are different: ResNet50 for RRT, SuperPoint~\cite{superpoint2018} for SuperGlue.
In this experiment, we use SuperPoint~\cite{superpoint2018} as the feature extractor for RRT, so that it has the same backbone architecture as SuperGlue.
We compare the new model with SuperGlue on Stanford Online Products~\cite{sop2016}.
We also explore finetuning the SuperPoint backbone (we tried finetuning SuperPoint on Google Landmarks v2-clean~\cite{gldv2} but found it requires much more computing resources than we can afford).
The SuperGlue model in this experiment is pretrained on ScanNet~\cite{scannet2017}. 
ScanNet is a large-scale dataset that contains 2.5 million images of 1513 indoor scenes.
Both SuperGlue and our method take a 320x320 grayscale image as input.
We extract the global descriptor by averaging all the local responses, and sample the top-500 local descriptors for all the models. 
We also investigate the benefit of using the position embedding for this task. 
The training and evaluation settings remain the same as in the SOP experiment presented in the main paper. 
We do not finetune SuperGlue on SOP as SOP does not include pixel-level annotations.

As shown in Table~\ref{tab:rrt_vs_glue_sop}, reranking by either SuperGlue or RRT can significantly improve the retrieval performance. 
RRT outperforms SuperGlue with a frozen SuperPoint backbone. 
Interestingly, RRT does not benefit from the position embedding in this task, as is also the case in the DELG experiment.
On the other hand, we observe that the position embedding is helpful in the SOP experiment of the main paper, where the descriptors of all the grid positions are used. 
We conjecture that the keypoints sampling may result in imbalanced sampled positions that potentially hinder the training. 
Finally, finetuning the SuperPoint backbone leads to the best performance.

\section{ResNet101 as the CNN backbone.}
\label{sec:r101}
\begin{table}[t]
\scalebox{0.73}{
\setlength{\tabcolsep}{3pt}
\centering
{
\begin{tabular}{l l c c c c c c c c}
\toprule
\multirow{2}{*}{Method}  & Desc. & \# local & Desc. & \multicolumn{2}{c}{ Medium} && \multicolumn{2}{c}{ Hard} \\
 & version & desc. & dim. & \multicolumn{1}{c}{\normalsize \vphantom{M} \normalsize $\mathcal{R}$Oxf } & \multicolumn{1}{c}{\normalsize $\mathcal{R}$Par} && \multicolumn{1}{c}{\normalsize $\mathcal{R}$Oxf} & \multicolumn{1}{c}{\normalsize $\mathcal{R}$Par} \\
\midrule
DELG global & R101-v1 & 0 & - & $73.2$ & $82.4$ && $51.2$ & $64.7$ \\
\midrule
GV & R101-v1 & 1000 & 128 & $78.5$ & $82.9$ && $59.3$ & $65.5$ \\
SuperGlue & SuperPoint & 500 & 256 & $74.6$ & $82.5$ && $51.7$ & $62.5$ \\
SuperGlue & SuperPoint & 1024 & 256 & $76.9$ & $82.9$ && $57.2$ & $64.7$ \\
RRT (ours) & R101-v1 & 500 & 128 & $\num[math-rm=\mathbf]{78.8}$ & $\num[math-rm=\mathbf]{83.2}$ && $\num[math-rm=\mathbf]{62.5}$ & $\num[math-rm=\mathbf]{68.4}$ \\
\midrule
DELG global & R101-v2-clean &0  &- & $76.3$ & $86.6$ && $55.6$ & $72.4$\\
\midrule
GV  & R101-v2-clean & 1000 & 128 & $\num[math-rm=\mathbf]{81.2}$ & $87.2$ && $64.0$ & $72.8$\\
SuperGlue  & SuperPoint & 500 & 256 & ${77.1}$ & $86.8$ && $55.5$ & $69.3$\\
SuperGlue  & SuperPoint & 1024 & 256 & ${79.7}$ & $87.1$ && $62.1$ & $71.5$\\
RRT (ours) & R101-v2-clean & 500 & 128 & $79.9$ & $\num[math-rm=\mathbf]{87.6}$ && $\num[math-rm=\mathbf]{64.1}$ & $\num[math-rm=\mathbf]{76.1}$ \\
\bottomrule
\end{tabular}
}
}
\caption{Comparison to geometric verification~\cite{spatial2007} and SuperGlue~\cite{superglue2020} on Revisited Oxford/Paris~\cite{revisited} using ResNet101~\cite{resnet2016} as the backbone. The SuperGlue model is pretrained on MegaDepth~\cite{megadepth2018} with SuperPoint~\cite{superpoint2018} as the backbone.
The mAP scores on the Medium and Hard setups are reported.}
\label{tab:rt_vs_gv_r101}
\end{table}
Following~\cite{delg2020}, we perform experiments using ResNet101 as the CNN backbone. 
We train the Reranking Transformer on two extra sets of image descriptors: the DELG R101 descriptors pretrained on Google Landmarks (GLD) v1~\cite{delf2017} and v2-clean~\cite{gldv2}. 
The training and evaluation settings remain the same as the main experiment on ResNet50, except that we also clip the gradient with a maximal norm of 0.1, and find that it can stabilize the training and lead to better performance. 
Here we compare our model with geometry verification~\cite{spatial2007} and SuperGlue~\cite{superglue2020} (pretrained on MegaDepth~\cite{megadepth2018}) on Revisited Oxford/Paris~\cite{revisited}, as shown in Table~\ref{tab:rt_vs_gv_r101}.

When evaluated on the ``v1'' descriptors, our method performs favorably to both geometry verification and SuperGlue on all the settings. 
When evaluated on the ``v2-clean'' descriptors, our method is inferior to geometry verification on $\mathcal{R}$Oxf-Medium but performs favorably to geometry verification and SuperGlue on the rest settings.

\section{Visualizing the correspondences}
\label{sec:correspondences}
\begin{figure}[t]
  \centering
  \includegraphics[width=0.85\textwidth]{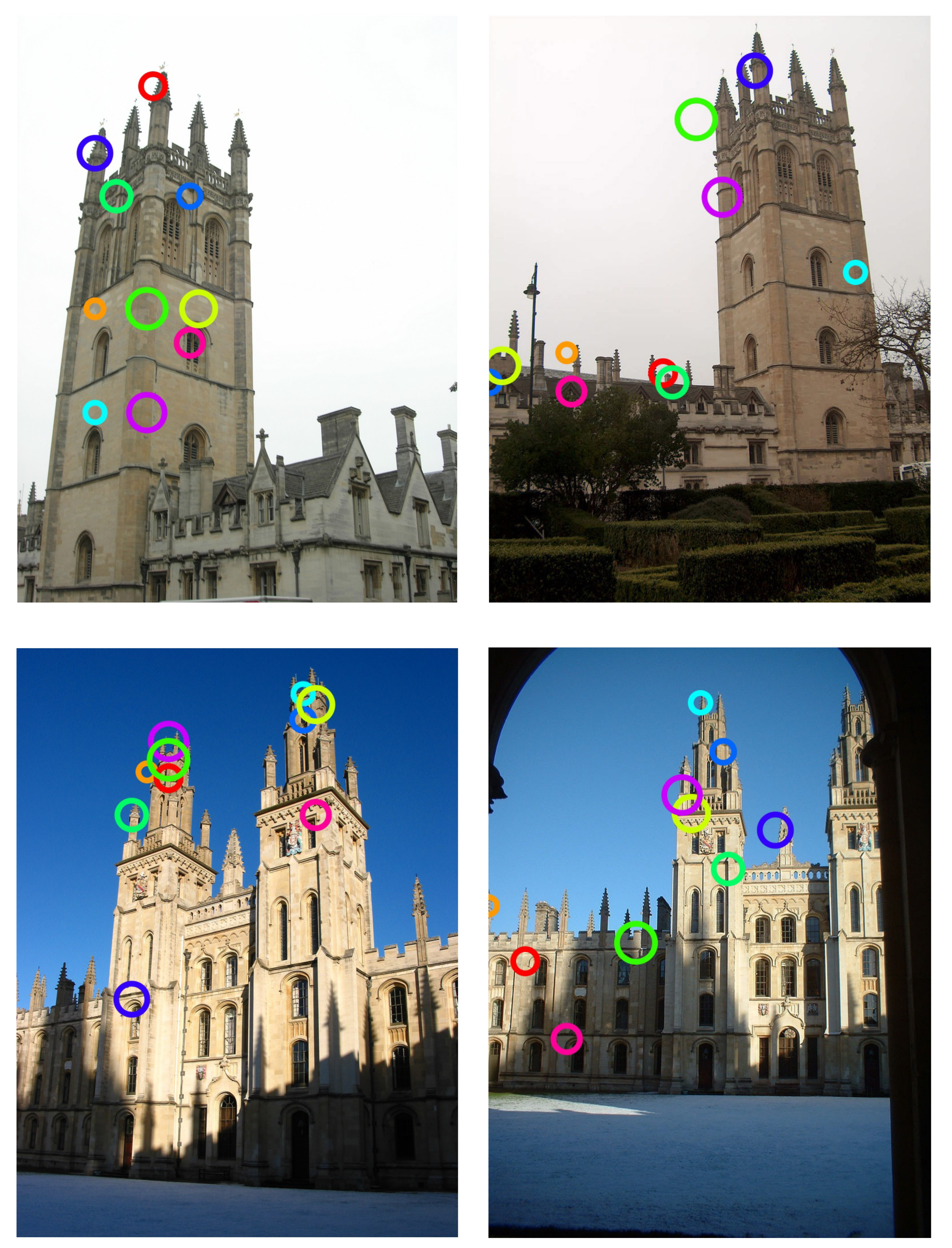}
  \caption{Visualization of the correspondences estimated by a trained RRT model (RRT-R50-v2-clean). 
  Each row shows a pair of matching images. 
  Two keypoints with the same color and scale are considered as a correspondence.}
  \label{fig:correspondences}
\end{figure}

Following the previous instance recognition~\cite{how2020} and image matching~\cite{superglue2020} works, we visualize the correspondences learned by RRT in Fig.~\ref{fig:correspondences}.
We extract the attention scores from the last transformer layer (i.e. $\mathbf{Z}_C$) of RRT.
Correspondences are computed by solving a linear sum assignment problem~\cite{assignment_problem} using the attentions as the affinity. 
The examples show that RRT is not good at learning the pixel-wise correspondences of keypoints. 
It also indicates that rather than estimating the local correspondences, RRT learns distinct knowledge to compute the similarity of images.

\section{Limitation}
\label{sec:limitation}
\textbf{Interpretability}.
Compared to the homography that explicitly models the alignment of the image-pair, the similarity score predicted by our model is less interpretable. In the future, we'd like to extend the work to learning more visual relation concepts, e.g. homography, dense matching, optical flow, which may lead to more interpretable results. 

\textbf{Domain shift}.
In the DELG~\cite{delg2020} experiment, our method is trained on Google Landmarks v2~\cite{gldv2} and tested on Revisited Oxford/Paris~\cite{revisited}.
In the SOP~\cite{sop2016} experiment, the training and test sets have no overlapping instance categories.
Both experiments demonstrate that the proposed Reranking Transformer can transfer the knowledge across different instance categories to a certain extent.
On the other hand, similar to all learning-based approaches, our method might have difficulty in handling large domain shifts.
It is also a major challenge for most of the recent approaches as another key component of the image retrieval pipeline, the feature extractor, may also suffer from domain shift. 
Learning transferable feature representation/matching could be an interesting topic for future research.

\section{More qualitative examples}
\label{sec:examples}
\begin{figure*}[h]
  \centering
  \includegraphics[width=0.99\textwidth]{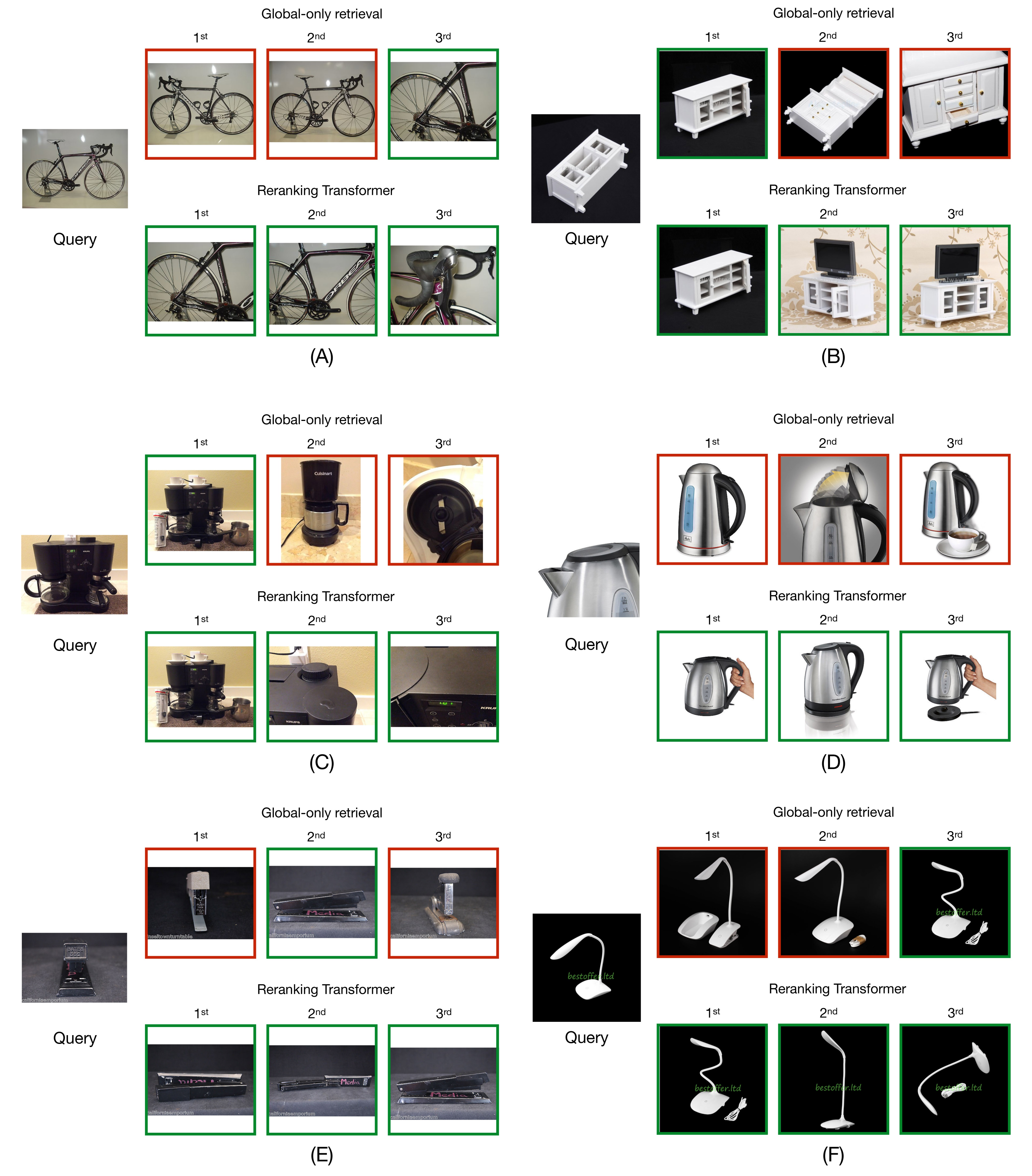}
  \caption{Qualitative examples from Stanford Online Products~\cite{sop2016}. For each query, the top-3 neighbors predicted by the global-only retrieval and the proposed Reranking Transformer are presented. Correct/incorrect neighbors are marked with {\color{green}green}/{\color{red}red} borders.}
  \label{fig:sop_rrt}
\end{figure*}

In Fig.~\ref{fig:sop_rrt}, we provide qualitative examples on Stanford Online Products~\cite{sop2016}.
Here, we compare the results from the global-only model (\textit{CO}) and the proposed model (\textit{CO + RRT (finetuned)}).
In particular, we showcase the examples of rigid objects (e.g. coffee maker, kettle) and deformable objects (e.g. stapler, lamp).
The proposed method outperforms the global-only retrieval on challenging cases such as partial-matching (example (A)(C)(D)), articulated objects (example (E)(F)), and irrelevant context (example (B)).

\begin{figure*}[h]
  \centering
  \includegraphics[width=0.99\textwidth]{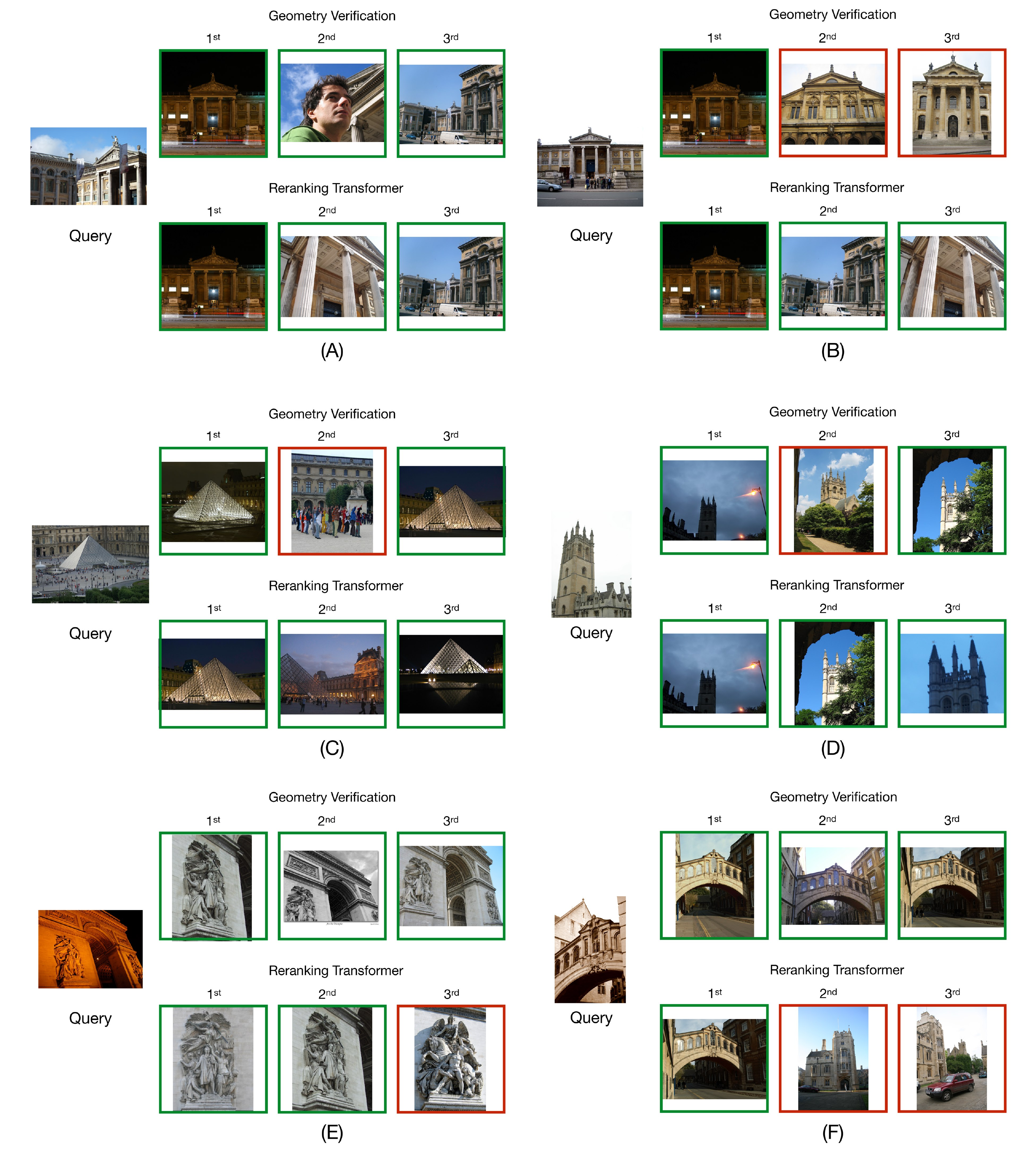}
  \caption{Qualitative examples from Revisited Oxford/Paris~\cite{revisited}. For each query, the top-3 neighbors predicted by geometry verification and the proposed Reranking Transformer are presented. Correct/incorrect neighbors are marked with {\color{green}green}/{\color{red}red} borders.}
  \label{fig:gv_rrt}
\end{figure*}

In Fig.~\ref{fig:gv_rrt}, we provide reranking examples produced by geometry verification and the proposed Reranking Transformer on Revisited Oxford/Paris~\cite{revisited}. 
It is shown that, compared to geometry verification, the proposed method performs favorably when large viewpoint variations are present. 
For example, the queries in example (A) and (B) represent the same landmark but exhibit a large viewpoint change. 
While geometry verification predicts two different sets of top neighbors, our model predicts the same set of top ranked images for the two queries. 
Example (E) and (F) show failure cases of our model.

\clearpage

\clearpage
{\small
\bibliographystyle{ieee_fullname}
\bibliography{egbib}
}

\end{document}